\documentclass[conference]{IEEEtran}

\usepackage{amsmath,amssymb,mathtools,bm}
\renewcommand{\bm}[1]{#1}
\usepackage{booktabs,array,multirow}
\usepackage{graphicx}
\usepackage{cite}
\usepackage{url}
\usepackage{tikz}
\usetikzlibrary{arrows.meta,positioning}

\newcommand{\R}{\mathbb{R}}
\newcommand{\Z}{\mathbb{Z}}

\newcommand{\argmaxop}{\operatorname*{arg\,max}}

\begin{document}

\title{Graph4BiLO: Graph Neural Network Approximation for Bilevel Mixed-Integer Linear Optimization}

\author{Jessica~D.~Elrefaei,
        Kaixun~Hua, Seungbae Kim,        Hoang~Nam~Tran,
        and~Juan~S.~Borrero%
\thanks{J. D. Elrefaei, H. N. Tran, and Seungbae Kim are with the Bellini College of Artificial Intelligence, Cybersecurity and Computing, University of South Florida. J. D. Elrefaei is also with the Muma College of Business, University of South Florida. K. Hua and J. S. Borrero are with the Department of Industrial and Management Systems Engineering, University of South Florida.}}

\maketitle

\begingroup
\renewcommand\thefootnote{}
\footnotetext{This work has been submitted to the IEEE for possible
publication. Copyright may be transferred without notice, after which this
version may no longer be accessible.}
\addtocounter{footnote}{-1}
\endgroup

\begin{abstract}
Bilevel mixed-integer linear optimization problems model hierarchical decision processes in which a leader anticipates the optimal response of a follower. Although expressive, these problems are computationally challenging because lower-level optimality is embedded in the leader's feasible region. Value-function reformulations replace the nested follower optimization with a constraint involving the follower's optimal value, but evaluating this value function exactly can itself be expensive. This paper introduces Graph4BiLO, a graph neural network (GNN) approach for learning bilevel value functions from variable--constraint graph representations. In contrast to fixed-length multilayer perceptron (MLP) representations, the GNN uses shared message-passing parameters and can therefore be applied across multiple problem sizes with a single trained model. The learned ReLU network is encoded exactly as mixed-integer linear constraints and embedded in an approximate single-level formulation. A repair step subsequently re-solves the follower problem for the selected leader decision to recover a bilevel-feasible follower response. We evaluate Graph4BiLO on knapsack interdiction instances with 20--100 items against the exact MibS solver and the learning-based Neur2BiLO method. Graph4BiLO obtains objective values comparable to Neur2BiLO across all tested sizes while avoiding size-specific neural networks. An additional
out-of-distribution experiment demonstrates zero-shot transfer from 20-item
training instances to previously unseen 40- and 60-item instances. However, embedding message passing at every graph node substantially increases the resulting mixed-integer formulation size and solve time. These results identify a central tradeoff between size-generalizable graph representations and the computational cost of embedding GNNs within optimization models.
\end{abstract}


\section{Introduction}
\IEEEPARstart{B}{ilevel} optimization models hierarchical decision making
between two optimization problems, where a leader acts while anticipating
the optimal response of a follower \cite{bard1998practical, dempe2002foundations}.
When integrality restrictions are present, bilevel mixed-integer linear
optimization problems (BMILPs) can represent interdiction, pricing, network
design, and other strategic decision problems, but the nested optimality
requirement makes them substantially more difficult than conventional
single-level mixed-integer linear programs \cite{vasquez2025single}.

A useful perspective is provided by the lower-level value function. For a fixed leader decision, the value function returns the follower's optimal objective value. This allows the nested follower optimization to be expressed through follower feasibility together with a value-function optimality constraint \cite{dempe2007new}. The resulting formulation is single-level in form, but it does not altogether eliminate the fundamental computational difficulty: evaluating the exact value function may require solving a mixed-integer optimization problem for each leader decision encountered during optimization \cite{vasquez2025single}.

Recent learning-based methods address this bottleneck by approximating the lower-level value function with a neural network. Neur2BiLO, for example, combines a permutation-invariant DeepSets representation with a multilayer perceptron (MLP) predictor to learn bilevel value functions \cite{dumouchelle2024neur2bilo}. Zhou et al. developed a learning-based approach for bilevel programs with
binary tender in which a MLP approximates the lower-level optimal
value as a function of the linking binary variables, and the learned value
function is then embedded in a single-level mixed-integer reformulation
\cite{zhou2024learning}. ReLU networks are particularly attractive in this setting because a trained network can be represented exactly using mixed-integer linear constraints and embedded directly in an optimization model \cite{fischetti2018deep, anderson2020strong}.

Although the set-based Neur2BiLO architecture is in principle compatible
with variable-sized instances, its reported knapsack-interdiction experiments
train a separate neural network for each problem size \cite{dumouchelle2024neur2bilo, neur2bilo_github}.
In contrast, we explicitly evaluate Graph4BiLO using a single model across
multiple problem sizes, including zero-shot transfer to previously unseen
sizes without retraining or fine-tuning. Optimization problems also possess
natural relational structure: variables participate in constraints through
coefficients, and the sparsity pattern of the constraint matrix defines a graph. GNNs provide a natural mechanism for exploiting this structure because their
local update rules share parameters across nodes and can operate on
variable-sized graphs. Recent work has further demonstrated that trained GNNs
can be formulated using mixed-integer programming and embedded directly within
larger optimization models \cite{zhang2024augmenting}.

Graph-based representations have consequently been used in a variety of
single-level combinatorial and mixed-integer optimization settings, including
learning branch-and-bound policies \cite{gasse2019exact},
GNN-guided predict-and-search methods \cite{han2023gnn}, and scalable primal heuristics
\cite{canturk2024scalable}. More recently, GNNs have also been applied to
bilevel interdiction problems. Kwon et al. developed a GNN-based heuristic
for the knapsack interdiction problem \cite{kwon2025deep}, while
Zhang et al. proposed a multipartite GNN framework for network interdiction
problems \cite{networkinterdictiongoesneural}. To our knowledge, however,
prior GNN-based bilevel approaches have not considered learning the
lower-level value function from a variable--constraint representation and
embedding the resulting GNN surrogate directly within a value-function
reformulation.

Knapsack interdiction itself has been studied extensively from both
complexity and exact-optimization perspectives. Caprara et al. analyze the
computational complexity of bilevel knapsack variants
\cite{caprara2014study} and develop exact methods for bilevel knapsack
problems with interdiction constraints
\cite{caprara2016bilevel}. Fischetti et al. further study
interdiction games and structural monotonicity properties with applications
to knapsack problems \cite{fischetti2019interdiction}.

This paper introduces \emph{Graph4BiLO}, a GNN-based value-function approximation framework for BMILPs. Graph4BiLO represents an optimization instance as a variable--constraint graph, trains a GNN to predict the lower-level value function from sampled leader decisions, encodes the trained ReLU network as mixed-integer linear constraints, and embeds the prediction in a single-level approximation. We study the framework using knapsack interdiction and compare it with MibS \cite{tahernejad2020mibs} and Neur2BiLO \cite{dumouchelle2024neur2bilo}.

The principal contributions are:
\begin{itemize}
    \item a variable--constraint graph representation for learning bilevel value functions;
    \item a GNN surrogate whose shared parameters permit one trained model to process multiple problem sizes;
    \item an end-to-end optimization framework that embeds the trained ReLU GNN in a mixed-integer approximation and repairs the resulting follower solution; and
    \item a computational study comparing Graph4BiLO with exact and
    learning-based baselines, evaluating zero-shot generalization to unseen
    problem sizes, and quantifying the predictive and computational tradeoffs
    associated with GNN depth and global information sharing.
\end{itemize}

\section{Bilevel Value-Function Formulation}

\subsection{Bilevel Mixed-Integer Linear Optimization}
Consider the linear BMILP

\begin{align*}
\min_{x,y}\quad
    & f^\top x + g^\top y \\
\text{s.t.}\quad
    & Ax+By\ge a, \label{eq:upper_con}\\
    & x\in\Z^{p_x}\times\R^{q_x}, \\
    & y\in
    \argmaxop_{y'}
    \left\{
        q^\top y':
        Cx+Dy'\ge b,\;
        y'\in\Z^{p_y}\times\R^{q_y}
    \right\},
\end{align*}
where all matrices and vectors are of appropriate dimensions. The leader selects $x$, while the follower responds with an optimal solution $y$. Thus, lower-level optimality is part of the leader's feasible region. Discrete bilevel linear optimization can be
\(\Sigma_2^P\)-hard, reflecting the additional complexity introduced by the
nested optimality requirement \cite{jeroslow1985polynomial}.

\subsection{Value-Function Reformulation}
Define the follower value function
\begin{equation*}
\phi(x)=
\max\left\{
q^\top y:
Cx+Dy\ge b,\;
y\in\Z^{p_y}\times\R^{q_y}
\right\}.
\label{eq:value_function}
\end{equation*}
For any follower-feasible $y$, $q^\top y\leq \phi(x)$ by definition of the maximum. Consequently, follower feasibility together with
\begin{equation*}
q^\top y\geq \phi(x)
\label{eq:value_constraint}
\end{equation*}
forces equality and therefore lower-level optimality. The bilevel model can thus be written as
\begin{align*}
\min_{x,y}\quad
    & f^\top x + g^\top y\\
\text{s.t.}\quad
    & Ax+By\ge a,\\
    & Cx+Dy\ge b,\\
    & q^\top y\ge\phi(x),\\
    & x\in\Z^{p_x}\times\R^{q_x},\\
    & y\in\Z^{p_y}\times\R^{q_y}.
\end{align*}
The nested optimization has been replaced by a value-function constraint, but
evaluating \(\phi(x)\) can itself require solving a difficult discrete
optimization problem. In the knapsack interdiction setting considered here,
each exact value-function evaluation requires solving a 0--1 knapsack problem,
a classical NP-hard combinatorial optimization problem
\cite{karp1972reducibility}.

\section{Graph4BiLO}

\subsection{Overview}
Graph4BiLO approximates the exact value function with a learned GNN surrogate. For a fixed problem class, random problem instances are generated and leader decisions $x^{(i)}$ are sampled. For each sample, the lower-level problem is solved exactly to obtain $\phi(x^{(i)})$. The corresponding optimization instance and sampled decision are represented as a graph $\mathcal{G}(x^{(i)})$, producing supervised graph--value pairs
\begin{equation*}
\left(
\mathcal{G}(x^{(i)}),
\phi(x^{(i)})
\right)_{i=1}^{N}.
\end{equation*}
A GNN $\phi_G(\cdot;\theta)$ is trained on these pairs. After training, $\theta$ is fixed, the ReLU network is encoded as mixed-integer linear constraints, and the surrogate is embedded into an approximate single-level optimization problem using established mixed-integer formulations for trained ReLU networks
\cite{fischetti2018deep,anderson2020strong}. Finally, the candidate leader decision is retained while the original follower problem is re-solved exactly to repair the follower solution.

\subsection{Bilevel Formulation}
We evaluate Graph4BiLO on knapsack interdiction. The leader interdicts items subject to an interdiction budget, while the follower packs the remaining items to maximize profit. The leader therefore seeks to minimize the follower's optimal achievable profit:
\begin{align*}
\min_{\bm{x}\in\{0,1\}^{n}}\quad
& \phi(\bm{x})\\
\text{s.t.}\quad
& \bm{v}^{\top}\bm{x}\le K,
\end{align*}
where the follower value function is
\begin{equation*}
\phi(\bm{x})
=
\max_{\bm{y}\in\{0,1\}^{n}}
\left\{
\bm{c}^{\top}\bm{y}
:
\bm{w}^{\top}\bm{y}\le d,\;
\bm{x}+\bm{y}\le\bm{1}
\right\}.
\label{eq:knapsack_phi}
\end{equation*}
Here, $c_i$, $w_i$, and $v_i$ denote item profit, knapsack weight, and interdiction cost, respectively; $d$ is the follower capacity and $K$ is the leader interdiction budget.

For fixed $\bm{x}$, the follower constraints can be written compactly as
\begin{equation*}
F\bm{y}+L\bm{x}\leq\bm{f},
\end{equation*}
where
\begin{equation*}
F=
\begin{bmatrix}
\bm{w}^{\top}\\
I
\end{bmatrix},
\qquad
L=
\begin{bmatrix}
\bm{0}^{\top}\\
I
\end{bmatrix},
\qquad
\bm{f}=
\begin{bmatrix}
d\\
\bm{1}
\end{bmatrix}.
\end{equation*}

\subsection{Variable--Constraint Graph Construction}
\label{sec:graph_construction}

A central idea in Graph4BiLO is to represent each bilevel instance as a graph
rather than as a fixed-length feature vector. For the knapsack-interdiction
formulation introduced above, the matrices \(F\) and \(L\) directly define
the interactions between follower variables, leader variables, and
constraints. Graph4BiLO uses these algebraic dependencies to construct a
heterogeneous variable--constraint graph. The graph contains three node types:
\begin{itemize}
    \item follower-variable nodes \(y_i\),
    \item leader-variable nodes \(x_k\), and
    \item constraint nodes \(c_j\).
\end{itemize}

The graph is constructed directly from the nonzero pattern of the matrices \(F\) and \(L\). Specifically:
\begin{enumerate}
    \item If \(F_{ji}\neq 0\), add a bidirectional edge between follower node \(y_i\) and constraint node \(c_j\) with edge weight \(F_{ji}\).
    \item If \(L_{jk}\neq 0\), add a bidirectional edge between leader node \(x_k\) and constraint node \(c_j\) with edge weight \(L_{jk}\).
    \item If \(F_{ji}\neq 0\) and \(L_{jk}\neq 0\) occur in the same constraint row \(j\), add a bidirectional \emph{shortcut edge} between \(y_i\) and \(x_k\). In the knapsack interdiction case, these shortcut edges are assigned unit weight.
\end{enumerate}

The variable--constraint representation is natural because an edge is created
precisely when a variable directly participates in a constraint. Thus, the
graph preserves the dependency structure of the underlying optimization
model rather than imposing an arbitrary notion of neighborhood. Message
passing can consequently be interpreted as propagating information along
the same interactions that determine feasibility and objective value in the
optimization problem, an idea that has also motivated variable--constraint
graph representations for single-level MILPs \cite{gasse2019exact}.

For example, in knapsack interdiction, the capacity constraint is connected
to every follower variable \(y_i\) through an edge weighted by \(w_i\).
These edges expose to the GNN both which items compete for the common
knapsack capacity and the strength of their contribution to that constraint.
In contrast, each linking constraint \(x_i+y_i\leq 1\) connects the leader
decision \(x_i\) directly to the corresponding follower decision \(y_i\),
capturing the key interdiction mechanism: selecting \(x_i=1\) prevents the
follower from selecting item \(i\). The additional \(x_i\)--\(y_i\) shortcut
edge makes this leader--follower interaction directly accessible during
message passing rather than requiring information to travel through the
intermediate constraint node.

The matrix-induced edges therefore preserve the sparse algebraic structure
of the follower problem, while the shortcut edges provide a direct connection
between leader and follower variables that co-occur in the same linking
constraint.

For knapsack interdiction, there is one capacity constraint and \(n\) linking constraints of the form \(x_i+y_i\le 1\). Consequently, a problem with \(n\) items yields
$n \text{ follower-variable nodes}$,
$n \text{ leader-variable nodes}$, and
$(n+1) \text{ constraint nodes}$,
for a total of \((3n+1)\) nodes.

\begin{figure}[t]
\centering
\resizebox{\columnwidth}{!}{%
\begin{tikzpicture}[
    scale=1.0,
    transform shape,
    node distance=0.78cm and 1.65cm,
    var/.style={
        circle,
        draw,
        minimum size=7mm,
        inner sep=1pt
    },
    con/.style={
        circle,
        draw,
        fill=gray!15,
        minimum size=7mm,
        inner sep=1pt
    },
    leader/.style={
        circle,
        draw,
        fill=blue!10,
        minimum size=7mm,
        inner sep=1pt
    },
    bidir/.style={
        {Latex[length=1.4mm]}-{Latex[length=1.4mm]},
        thin
    },
    shortcut/.style={
        {Latex[length=1.4mm]}-{Latex[length=1.4mm]},
        thin,
        dashed,
        blue!70
    }
]

\node[var] (y1) {$y_1$};
\node[var, below=of y1] (y2) {$y_2$};
\node[var, below=of y2] (y3) {$y_3$};

\node[con, right=2.0cm of y1] (c1) {$c_1$};
\node[con, below=of c1] (c2) {$c_2$};
\node[con, below=of c2] (c3) {$c_3$};
\node[con, below=of c3] (c4) {$c_4$};

\node[leader, right=2.0cm of c1] (x1) {$x_1$};
\node[leader, below=1.18cm of x1] (x2) {$x_2$};
\node[leader, below=1.18cm of x2] (x3) {$x_3$};

\draw[bidir]
    (y1) --
    node[pos=0.52,above,sloped] {\scriptsize $w_1$}
    (c1);

\draw[bidir]
    (y2) --
    node[pos=0.38,above=2pt,sloped] {\scriptsize $w_2$}
    (c1);

\draw[bidir]
    (y3) --
    node[pos=0.48,below,sloped] {\scriptsize $w_3$}
    (c1);

\draw[bidir]
    (y1) --
    node[pos=0.68,above=2pt] {\scriptsize $1$}
    (c2);

\draw[bidir]
    (y2) --
    node[pos=0.55,above] {\scriptsize $1$}
    (c3);

\draw[bidir]
    (y3) --
    node[pos=0.55,above] {\scriptsize $1$}
    (c4);

\draw[bidir]
    (c2) --
    node[pos=0.55,above] {\scriptsize $1$}
    (x1);

\draw[bidir]
    (c3) --
    node[pos=0.55,above] {\scriptsize $1$}
    (x2);

\draw[bidir]
    (c4) --
    node[pos=0.55,above] {\scriptsize $1$}
    (x3);

\draw[shortcut,bend left=30]
    (y1) to
    node[pos=0.50,above,sloped] {\scriptsize $1$}
    (x1);

\draw[shortcut,bend left=30]
    (y2) to
    node[pos=0.50,above,sloped] {\scriptsize $1$}
    (x2);

\draw[shortcut,bend left=30]
    (y3) to
    node[pos=0.50,above,sloped] {\scriptsize $1$}
    (x3);

\node[anchor=west,font=\scriptsize] at (0.15,-5.40) {
    \tikz[baseline=-0.5ex]{\draw[bidir] (0,0)--(0.65,0);}
    \; Matrix-induced edge
};

\node[anchor=west,font=\scriptsize] at (3.80,-5.40) {
    \tikz[baseline=-0.5ex]{\draw[shortcut] (0,0)--(0.65,0);}
    \; Shortcut edge
};

\end{tikzpicture}%
}
\caption{Variable--constraint graph representation for a three-item knapsack interdiction instance. Follower-variable nodes are shown on the left, constraint nodes in the center, and leader-variable nodes on the right. Solid edges are induced directly from the matrices \(F\) and \(L\), while dashed edges are shortcut edges between leader and follower variables that co-occur in the same constraint row.}
\label{fig:vcg_knapsack}
\end{figure}
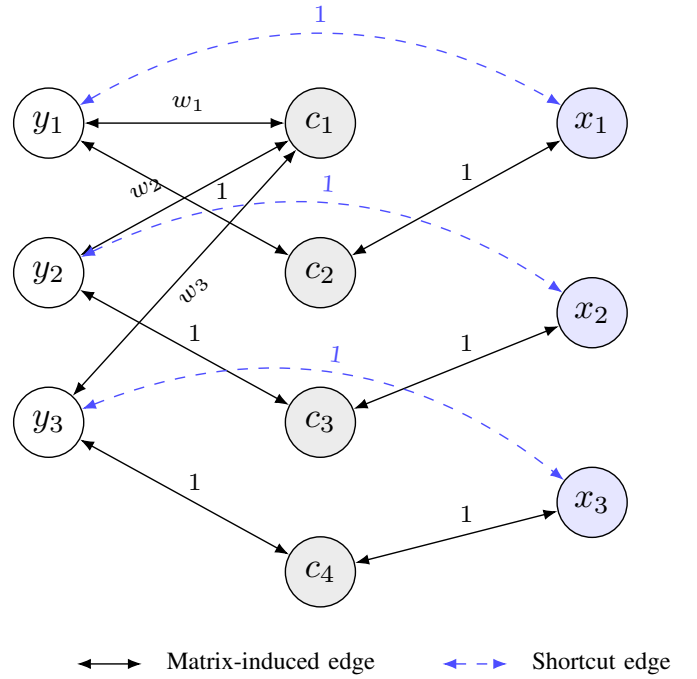

Fig.~\ref{fig:vcg_knapsack} illustrates the resulting graph representation for a three-item knapsack interdiction instance. The node \(c_1\) corresponds to the knapsack capacity constraint, while the nodes \(c_2,c_3,c_4\) correspond to the linking constraints \(x_i+y_i\le 1\). The weights \(w_1,w_2,w_3\) label the capacity-constraint edges, and the unit-weight edges reflect the coefficients in the linking constraints.

\subsection{Heterogeneous GNN Architecture}
\label{sec:heterogeneous_gnn}

Because the variable--constraint representation contains distinct leader-variable,
follower-variable, and constraint node types, Graph4BiLO uses a heterogeneous
message-passing architecture rather than a single homogeneous graph convolution.
Heterogeneous GNNs allow message and update functions to depend on node and
edge type, and PyTorch Geometric provides the \texttt{HeteroConv} abstraction
for assigning separate graph-convolution operators to different relation types
\cite{fey2019pytorch}. This construction is closely related to relational GNN
formulations in which distinct edge relations are assigned separate learned
transformations \cite{schlichtkrull2018modeling}. Graph4BiLO adopts this
principle using six directed relation types,
\[
\mathcal{R}
=
\{
y\!\to\!c,\;
c\!\to\!y,\;
x\!\to\!c,\;
c\!\to\!x,\;
x\!\to\!y,\;
y\!\to\!x
\}.
\]
Thus, leader-variable ($x$), follower-variable ($y$), and constraint nodes ($c$) all participate
in message passing and are updated at every GNN layer.

Conceptually, the architecture performs three steps. First, heterogeneous node
features are mapped into a common hidden space so that information associated
with leader decisions, follower decisions, and constraints can interact.
Second, relation-specific message passing propagates information along the
algebraic dependencies encoded by the optimization model. Finally, a
pooling-and-broadcast operation supplies each node with graph-level context
before the variable-sized collection of node representations is reduced to a
fixed-dimensional graph embedding for value prediction.

Let \(t\in\{y,x,c\}\) index follower-variable, leader-variable, and
constraint node types, with \(n_t\) nodes and feature matrix
\(X_t\in\mathbb{R}^{n_t\times f}\). Each node type is independently
projected into a common \(d\)-dimensional hidden space:
\[
H_t^{(0)}
=
X_tW_t^{\mathrm{in}}
+
\mathbf{1}_{n_t}(b_t^{\mathrm{in}})^\top,
\qquad t\in\{y,x,c\}.
\]
The type-specific projections place semantically different nodes in a common
hidden space while retaining separate learned transformations for each type.

Unlike a homogeneous GNN, Graph4BiLO assigns separate learned parameters to
each directed relation. Separate transformations are used because messages
between different node types have different optimization meanings. For
example, a constraint-to-variable message communicates information about the
restrictions acting on a decision, whereas a leader-to-follower message
communicates the direct effect of interdiction. Because these transformations are learned independently, different relations
can encode effects in different directions; for example, greater interdiction
can be associated with lower follower profit, whereas selecting more follower
items can be associated with higher follower profit. Let \(A_{t\rightarrow s}\) denote the weighted adjacency
matrix associated with messages sent from source node type \(t\) to destination
node type \(s\). For layer \(\ell\), the relation-specific graph convolution can
be written as
\begin{align*}
M_{t\rightarrow s}^{(\ell)}
={}&
H_s^{(\ell-1)}
W_{\mathrm{self},\,t\rightarrow s}^{(\ell)}
\\
&+
A_{t\rightarrow s}
H_t^{(\ell-1)}
W_{\mathrm{nbr},\,t\rightarrow s}^{(\ell)}
\\
&+
\mathbf{1}_{n_s}
\left(
b_{\mathrm{nbr},\,t\rightarrow s}^{(\ell)}
\right)^\top,
\end{align*}
where
\(W_{\mathrm{self},\,t\rightarrow s}^{(\ell)}\) and
\(W_{\mathrm{nbr},\,t\rightarrow s}^{(\ell)}\)
are relation-specific \(d\times d\) weight matrices and
\(b_{\mathrm{nbr},\,t\rightarrow s}^{(\ell)}\in\mathbb{R}^{d}\).

For each destination type, incoming relation-specific messages are summed and
passed through a ReLU:
\[
H_s^{(\ell)}
=
\mathrm{ReLU}\!\left(
\sum_{t:(t\rightarrow s)\in\mathcal R}
M_{t\rightarrow s}^{(\ell)}
\right),
\qquad s\in\{y,x,c\}.
\]

Local message passing only communicates information within the receptive field
created by the chosen number of GNN layers. However, the follower value
\(\phi(x)\) is a graph-level quantity that may depend on interactions across
the complete optimization instance. Graph4BiLO therefore supplements local
message passing with a global pooling-and-broadcast operation, providing every
node with a summary of the entire instance without requiring additional local
message-passing layers. Therefore, after the final message-passing layer, Graph4BiLO applies a global
pooling-and-broadcast operation. Mean pooling is first performed separately
over the three node types:
\begin{align*}
\bar{h}_y
&=
\frac{1}{n_y}
\sum_{i=1}^{n_y}
H_y^{(L)}[i,:],\\
\bar{h}_x
&=
\frac{1}{n_x}
\sum_{i=1}^{n_x}
H_x^{(L)}[i,:],\\
\bar{h}_c
&=
\frac{1}{n_c}
\sum_{i=1}^{n_c}
H_c^{(L)}[i,:].
\end{align*}
These representations are concatenated to form a global graph summary
\[
g
=
[
\bar{h}_y
\mathbin{\|}
\bar{h}_x
\mathbin{\|}
\bar{h}_c
]
\in\mathbb{R}^{3d},
\]
where \(\mathbin{\|}\) denotes concatenation.

The global representation is then broadcast back to every node and combined
with the node's local representation through a type-specific learned
transformation:
\[
\widetilde{H}_t
=
\mathrm{ReLU}
\left(
[
H_t^{(L)}
\mathbin{\|}
\mathbf{1}_{n_t}g
]
W_t^{\mathrm{br}}
+
\mathbf{1}_{n_t}
\left(b_t^{\mathrm{br}}\right)^\top
\right),
\qquad
t\in\{y,x,c\}.
\]
Thus, each node retains its locally propagated representation while also
receiving graph-level context. Separate broadcast transformations are learned
for follower-variable, leader-variable, and constraint nodes.

The broadcast-enhanced node embeddings are then mean pooled separately by type:
\begin{align*}
\widetilde{h}_y
&=
\frac{1}{n_y}
\sum_{i=1}^{n_y}
\widetilde{H}_y[i,:],\\
\widetilde{h}_x
&=
\frac{1}{n_x}
\sum_{i=1}^{n_x}
\widetilde{H}_x[i,:],\\
\widetilde{h}_c
&=
\frac{1}{n_c}
\sum_{i=1}^{n_c}
\widetilde{H}_c[i,:].
\end{align*}
The pooled representations are concatenated to form
\[
h_G
=
[
\widetilde{h}_y
\mathbin{\|}
\widetilde{h}_x
\mathbin{\|}
\widetilde{h}_c
]
\in\mathbb{R}^{3d}.
\]

This second pooling step converts the variable number of node representations
into a fixed-dimensional graph representation. Consequently, the dimensionality
of the final readout does not depend on the number of items in the optimization
instance.

The structural graph representation is supplemented with two problem-level
reference features that provide additional global information about the scale
of the follower objective:
\[
z
=
[
h_G
\mathbin{\|}
r_1
\mathbin{\|}
r_2
]
\in\mathbb{R}^{3d+2}.
\]
Here, \(r_1\) is the follower objective obtained for the reference case with
no interdiction, \(x=0\), while \(r_2\) is the additional \(1.25\times K\) objective
feature used by the model, where \(K\) denotes the interdiction budget.

Finally, a linear readout produces the predicted lower-level value:
\[
\hat{\phi}
=
w^\top z+b,
\qquad
w\in\mathbb{R}^{3d+2},
\quad
b\in\mathbb{R}.
\]
The resulting prediction defines the learned value-function approximation
\(\phi_G(x;\theta)\).

The heterogeneous formulation allows each semantic relation in the
optimization graph to learn a distinct transformation while sharing the
parameters associated with that relation across all nodes and problem
instances. Because these operations do not depend on a fixed number of nodes,
the same trained Graph4BiLO model can be applied to graphs with different
numbers of variables and constraints.

\subsection{Training Objective}

The supervised training set is
\begin{equation*}
\mathcal{D}
=
\left\{
\left(\mathcal{G}^{(i)},x^{(i)},\phi(x^{(i)})\right)
\right\}_{i=1}^{N}.
\end{equation*}
The model is trained by minimizing the Huber loss between the predicted and
exact lower-level objective values. Let
\begin{equation*}
e_i
=
\phi_G(\mathcal{G}^{(i)},x^{(i)};\theta)
-
\phi(x^{(i)}).
\end{equation*}
Using threshold \(\delta=2\), the sample-wise loss is
\begin{equation*}
\mathcal{L}_{2}(e_i)
=
\begin{cases}
\frac{1}{2}e_i^2,
& |e_i|\leq 2,\\[4pt]
2\left(|e_i|-1\right),
& |e_i|>2.
\end{cases}
\end{equation*}
The training objective is therefore
\begin{equation*}
\min_{\theta}
\frac{1}{N}
\sum_{i=1}^{N}
\mathcal{L}_{2}\!\left(
\phi_G(\mathcal{G}^{(i)},x^{(i)};\theta)
-
\phi(x^{(i)})
\right).
\end{equation*}
The Huber loss behaves quadratically for small prediction errors and linearly
for larger errors, reducing the influence of large residuals compared with
mean squared error while still encouraging accurate value-function prediction.

\section{Embedding the Learned Value Function}

\subsection{Mixed-Integer ReLU Encoding}
Because the network uses ReLU activations, it can be embedded exactly in a mixed-integer linear model once valid pre-activation bounds are available \cite{fischetti2018deep, anderson2020strong}. For a single neuron,
\begin{equation*}
z=Wx+b,\qquad h=\max\{0,z\},
\end{equation*}
with $L\leq z\leq U$. Introducing a binary activation indicator $\delta$ yields
\begin{align*}
h &\ge z,\\
h &\ge 0,\\
h &\le z-L(1-\delta),\\
h &\le U\delta,\\
\delta &\in\{0,1\}.
\end{align*}
Applying these constraints to every hidden ReLU gives an exact MILP representation of the trained neural network.

\subsection{Approximate Single-Level Knapsack Model}
Graph4BiLO replaces the exact follower value with the learned approximation,
\begin{equation*}
\phi(\bm{x})\approx\phi_G(\bm{x};\theta).
\end{equation*}
For knapsack interdiction, the approximate single-level model is
\begin{align*}
\min_{\bm{x},\bm{y},s\geq0}\quad
& \bm{c}^{\top}\bm{y}+\lambda s\\
\text{s.t.}\quad
& \bm{v}^{\top}\bm{x}\le K,\\
& \bm{w}^{\top}\bm{y}\le d,\\
& \bm{x}+\bm{y}\le\bm{1},\\
& \bm{c}^{\top}\bm{y}
\ge
\phi_G(\bm{x};\theta)-s,\\
& \bm{x},\bm{y}\in\{0,1\}^{n}.
\label{eq:approx_model}
\end{align*}
The trained parameters $\theta$ are fixed during optimization. The nonnegative slack variable $s$ protects the approximate formulation against infeasibility caused by value-function overprediction, while $\lambda$ penalizes the relaxation in the objective.

\subsection{Repair and Bilevel Feasibility}
The approximate model returns a candidate $(\hat{\bm{x}},\hat{\bm{y}})$. Graph4BiLO keeps the leader decision fixed and re-solves the original follower problem:
\begin{equation*}
\bm{y}^{r}
\in
\argmaxop_{\bm{y}\in\{0,1\}^{n}}
\left\{
\bm{c}^{\top}\bm{y}:
\bm{w}^{\top}\bm{y}\le d,\;
\hat{\bm{x}}+\bm{y}\le\bm{1}
\right\}.
\end{equation*}
The exact follower value is then
\begin{equation*}
\phi(\hat{\bm{x}})=\bm{c}^{\top}\bm{y}^{r}.
\end{equation*}
Replacing $\hat{\bm{y}}$ with $\bm{y}^{r}$ restores lower-level optimality. The optimality residual
\begin{equation*}
\phi(\hat{\bm{x}})-\bm{c}^{\top}\bm{y}^{r}=0
\end{equation*}
therefore verifies that the repaired follower decision is optimal for the selected leader decision.

\section{Computational Experiments}

\subsection{Experimental Setup}
The study uses 10,000 generated knapsack interdiction instances: 2,000 instances for each problem size $n\in\{20,40,60,80,100\}$. One leader decision is sampled per instance, and the corresponding ground-truth value is obtained by solving the follower knapsack problem exactly. The data are divided into 80\% training, 10\% validation, and 10\% test sets using a stratified split.

The GNN uses two message-passing layers with hidden dimension 32, mean pooling, and a linear readout. Training uses AdamW with learning rate $10^{-3}$ for up to 500 epochs.

We compare Graph4BiLO with two baselines. MibS is an exact branch-and-cut solver for mixed-integer bilevel linear optimization \cite{tahernejad2020mibs}. Neur2BiLO is a learning-based value-function approximation method that combines a permutation-invariant DeepSets representation with an MLP predictor \cite{dumouchelle2024neur2bilo}. For each problem size, all three methods are evaluated on the same 10
knapsack-interdiction test instances, with a maximum optimization time of
3600 seconds per instance.

\subsection{Computational Performance}
Table~\ref{tab:performance} reports mean objective values and solve times. Because the leader minimizes the follower's optimal profit, lower objective values are preferred.

\begin{table*}[!t]
\caption{Mean Computational Performance on Knapsack Interdiction}
\label{tab:performance}
\centering
\renewcommand{\arraystretch}{1.15}
\setlength{\tabcolsep}{7pt}
\begin{tabular}{c|cc|cc|cc}
\toprule
\multirow{2}{*}{$n$} &
\multicolumn{2}{c|}{Graph4BiLO} &
\multicolumn{2}{c|}{MibS} &
\multicolumn{2}{c}{Neur2BiLO}\\
& Obj. & Time (s) & Obj. & Time (s) & Obj. & Time (s)\\
\midrule
20  & 0.58 & 29.52   & 0.55 & 2864.89 & 0.54 & 0.11\\
40  & 0.99 & 1049.44 & 1.81 & 3600.00 & 0.97 & 0.06\\
60  & 1.66 & 3600.21 & 3.62 & 3600.00 & 1.65 & 0.08\\
80  & 2.10 & 3600.19 & 4.89 & 3600.00 & 2.07 & 0.11\\
100 & 2.66 & 3600.34 & 6.68 & 3600.00 & 2.62 & 0.15\\
\midrule
Overall & 1.60 & 2375.94 & 3.63 & 3452.98 & 1.57 & 0.10\\
\bottomrule
\end{tabular}
\end{table*}

Graph4BiLO achieves mean objective values close to Neur2BiLO across all tested sizes. From $n=40$ onward, both learning-based approaches return substantially lower mean objectives than MibS within the imposed time limit. Graph4BiLO and Neur2BiLO remain especially close: their mean objectives differ by 0.02 at $n=40$, 0.01 at $n=60$, 0.03 at $n=80$, and 0.04 at $n=100$.

At \(n=20\), Graph4BiLO obtains a slightly higher mean objective
(\(0.58\)) than MibS (\(0.55\)) and Neur2BiLO (\(0.54\)). This small gap is
consistent with the relative advantages of the competing methods at the
smallest problem size. MibS can solve the smaller bilevel instances more
effectively within the time limit, while the reported Neur2BiLO benchmark
uses a separately trained model for each problem size
\cite{dumouchelle2024neur2bilo,neur2bilo_github}. Graph4BiLO instead uses a
single set of learned parameters across all five sizes. Thus, Neur2BiLO
benefits from size-specific specialization in this comparison, whereas
Graph4BiLO trades some specialization for cross-size parameter sharing and
generalization.

The principal difference is computational time. Neur2BiLO solves the approximate optimization problem in approximately 0.1 seconds on average, whereas Graph4BiLO requires 29.52 seconds at $n=20$, 1049.44 seconds at $n=40$, and reaches the one-hour limit for $n\geq60$. Thus, the graph representation provides cross-size parameter sharing but produces a substantially larger embedded optimization model.

\subsection{Generalization to Unseen Problem Sizes}
\label{sec:ood}

A principal advantage of the graph-based representation is that the same
trained model can be applied to instances containing different numbers of
variables and constraints. To evaluate this capability directly, we conduct
an out-of-distribution (OOD) size experiment in which Graph4BiLO is trained
exclusively on 2,000 knapsack-interdiction instances of size \(n=20\) and is
then applied, without retraining or fine-tuning, to previously unseen
instances of sizes \(n=40\) and \(n=60\). Thus, no labeled lower-level
training samples from either OOD size are used to train the model. Because the reported Neur2BiLO knapsack-interdiction experiments use
size-specific training and checkpoints \cite{dumouchelle2024neur2bilo},
we do not include it as a benchmark in this experiment, which specifically
evaluates zero-shot transfer of a single trained model to unseen problem sizes.

This setting is particularly relevant when supervised label generation is
computationally expensive. Each Graph4BiLO training target requires solving
the lower-level optimization problem to obtain the exact value
\(\phi(x)\). Under a size-specific training protocol, generating a new model for each
problem dimension also requires generating a corresponding set of exact
lower-level training labels. The ability to reuse a single trained model
across sizes can therefore reduce this repeated data-generation cost,
particularly when the lower-level problem is itself computationally expensive. In contrast, Graph4BiLO shares its
message-passing parameters across variable-sized graphs, allowing the cost of
training-data generation and model fitting to be amortized across multiple
problem sizes.

\begin{table*}[t]
\centering
\caption{Out-of-distribution generalization from \(n=20\) training instances
to unseen knapsack-interdiction sizes. Graph4BiLO is evaluated without
retraining or fine-tuning.}
\label{tab:ood_results}
\begin{tabular}{c l cc}
\toprule
OOD Size &
Method &
Mean Objective &
Mean Solve Time (s) \\
\midrule
40 &
MibS &
1.895 &
3600 \\

40 &
Graph4BiLO, $\lambda=0.1$ &
0.978 &
2.92 \\
\midrule
60 &
MibS &
3.489 &
3600 \\

60 &
Graph4BiLO, $\lambda=0.1$ &
1.609 &
5.81 \\
\bottomrule
\end{tabular}
\end{table*}

The results in Table~\ref{tab:ood_results} demonstrate that a Graph4BiLO
model trained only on \(n=20\) instances can be embedded and solved directly
on substantially larger problem sizes without generating a new training set
or retraining the network. At \(n=40\), the OOD Graph4BiLO model obtains a
mean leader objective of \(0.978\) with a mean solve time of only
\(2.92\) seconds. In comparison, MibS reaches the 3,600-second time limit
and reports a mean objective of \(1.895\).

The same pattern persists at \(n=60\), where Graph4BiLO obtains a mean
objective of \(1.609\) in \(5.81\) seconds, while MibS reaches the
3,600-second time limit with a mean objective of \(3.489\). Thus, the
zero-shot model remains effective even when the problem size triples from
the \(n=20\) instances used for training to \(n=60\) at test time.

More broadly, the experiment illustrates a practical setting in which the
variable-sized graph representation can provide an advantage over
size-specific neural surrogates. Training independent models for
\(n=20\), \(40\), and \(60\) using 2,000 samples per size would require
6,000 labeled lower-level solves in total, in addition to three separate
training procedures. The OOD Graph4BiLO experiment instead uses only the
2,000 labeled \(n=20\) samples and a single trained model. Although predictive
and optimization performance may degrade as the test distribution moves
farther from the training size, the ability to obtain solutions for unseen
problem dimensions without additional label generation or retraining can be
particularly valuable when exact lower-level solves are expensive, such as routing problems or traveling salesman problems \cite{papadimitriou1977euclidean, lenstra1981complexity}.

\subsection{Architecture Ablation}
\label{sec:architecture_ablation}

To examine the effect of message-passing depth on both predictive accuracy
and the computational cost of the embedded formulation, we conduct an
ablation study using knapsack-interdiction instances of size \(n=20\).
The hidden dimension is fixed at \(d=4\), while the number of
message-passing layers is varied from one to four. All architectures are
trained and evaluated using the same data-generation procedure and
train--test setting. We additionally evaluate the four-layer architecture
without the global pooling-and-broadcast operation to isolate the effect of
this component. Because the ablation models use a much smaller hidden
dimension than the main Graph4BiLO configuration (\(d=4\) versus \(d=32\)),
their absolute solve times are substantially lower and should be interpreted
primarily as relative comparisons among ablation settings rather than as
directly comparable to the runtimes in Table~\ref{tab:performance}.

\begin{table*}[t]
\centering
\caption{Architecture ablation on \(n=20\) knapsack-interdiction instances
with hidden dimension \(d=4\). Lower values are preferred for all reported
metrics, including the leader objective.}
\label{tab:architecture_ablation}
\begin{tabular}{lccccc}
\toprule
Architecture &
Huber Loss &
MAE &
RMSE &
Mean Objective &
Mean Solve Time (s) \\
\midrule
1 layer, $d=4$
& $1.157\times10^{-2}$
& $1.167\times10^{-1}$
& $1.521\times10^{-1}$
& \textbf{0.540}
& \textbf{0.280} \\

2 layers, $d=4$
& $\mathbf{1.013\times10^{-2}}$
& $\mathbf{1.148\times10^{-1}}$
& $\mathbf{1.424\times10^{-1}}$
& 0.600
& 3.200 \\

3 layers, $d=4$
& $1.138\times10^{-2}$
& $1.192\times10^{-1}$
& $1.508\times10^{-1}$
& 0.651
& 4.300 \\

4 layers, $d=4$
& $1.068\times10^{-2}$
& $1.184\times10^{-1}$
& $1.461\times10^{-1}$
& 0.577
& 3.490 \\

4 layers, $d=4$, no broadcast
& $1.449\times10^{-2}$
& $1.301\times10^{-1}$
& $1.703\times10^{-1}$
& 0.706
& 1.420 \\
\bottomrule
\end{tabular}
\end{table*}

Table~\ref{tab:architecture_ablation} shows that increasing
message-passing depth does not produce a monotonic improvement in predictive
accuracy. The two-layer model achieves the lowest Huber loss, MAE, and RMSE,
whereas the three- and four-layer models do not improve upon these values.
Moreover, the additional message-passing layers substantially increase the
computational burden of the embedded formulation. The one-layer model solves
in only \(0.28\) seconds on average, compared with \(3.20\), \(4.30\), and
\(3.49\) seconds for the two-, three-, and four-layer architectures,
respectively.

Interestingly, prediction accuracy and optimization performance are not
perfectly aligned. Although the two-layer model provides the lowest prediction
errors, the one-layer architecture produces the lowest mean leader objective,
\(0.5395\), followed by the four-layer model at \(0.5767\). This suggests
that small differences in value-function prediction error do not necessarily
translate directly into corresponding differences in the leader decisions
selected by the embedded optimization model.

The broadcast ablation further illustrates the tradeoff between predictive
quality and computational cost. Removing the global broadcast from the
four-layer model reduces mean solve time from \(3.49\) to \(1.42\) seconds,
but increases the Huber loss from \(0.010679\) to \(0.014493\), the MAE from
\(0.118443\) to \(0.130136\), and the mean leader objective from \(0.5767\)
to \(0.7060\). Thus, the global broadcast appears to improve the quality of
the learned value-function representation and the resulting optimization
solution, although this benefit comes at the cost of a more difficult
embedded formulation.

Overall, these results provide little evidence that increasing local
message-passing depth beyond two layers improves value-function prediction
for this benchmark. This is particularly relevant because deeper networks
not only introduce additional ReLU variables and constraints into the
embedded MILP, but also require bounds to be propagated through more
successive neural-network layers. Together with the representation-similarity
analysis in Section~\ref{sec:oversmoothing}, the ablation suggests that the
potential benefit of a larger local receptive field must be balanced against
both representation homogenization and increased optimization complexity.

\subsection{Embedded Formulation Size}

The formulation-size increase follows directly from the GNN architecture.
For a knapsack-interdiction instance with \(n\) items, the graph contains
\[
|V|=3n+1
\]
nodes. For \(n=100\), this gives
\[
|V|=301.
\]

Each message-passing layer produces a \(d\)-dimensional ReLU representation
for every graph node. In addition, the global pooling-and-broadcast operation
applies a final \(d\)-dimensional ReLU transformation to every node. The global broadcast therefore leaves the graph topology unchanged but adds
another node-wise nonlinear transformation to the embedded network, providing
a direct explanation for the solve-time increase observed in the broadcast
ablation below (Table~\ref{tab:architecture_ablation}). Thus,
with \(L\) message-passing layers and hidden dimension \(d\), the number of
node-wise hidden ReLU activations is approximately
\[
|V|d(L+1),
\]
where the additional term accounts for the post-message-passing broadcast
transformation.

For the architecture used in the main experiments, \(d=32\) and \(L=2\).
At \(n=100\), this gives
\[
301(32)(3)=28{,}896
\]
embedded ReLU activations. Without the broadcast transformation, the
corresponding count would be
\[
301(32)(2)=19{,}264.
\]

Each ambiguous ReLU may require a binary activation indicator together with
continuous pre- and post-activation variables. The embedded GNN can therefore
introduce up to approximately \(28{,}896\) binary variables and \(57{,}792\)
continuous variables at \(n=100\). Under the standard big-\(M\) encoding,
counting one affine relation together with four ReLU inequalities per hidden
activation gives up to approximately
\[
28{,}896(5)=144{,}480
\]
affine and activation constraints, in addition to the pooling, broadcast,
readout, graph-specific relations, and original optimization constraints.

\subsection{Tradeoff Between Receptive Coverage and GNN Depth}
\label{sec:oversmoothing}

A fundamental challenge for local message-passing GNNs is that information
can travel only one graph hop per layer. This is particularly important for
optimization problems because the quantity being predicted may depend on
interactions across the entire optimization instance rather than only on a
node's immediate neighborhood. Ideally, the learned representation should
therefore allow information from all relevant variables and constraints to
influence the prediction.

For the knapsack-interdiction graph used in Graph4BiLO, the graph diameter is
four. Consequently, four local message-passing layers are sufficient, and for
some node pairs necessary, for information to propagate between arbitrary
nodes through graph edges. A four-layer model therefore represents the depth
at which every node can, in principle, receive information originating from
the entire graph through local message passing alone.

Increasing the number of layers to obtain this full receptive coverage,
however, introduces two competing effects. First, repeated neighborhood
aggregation can cause initially distinct node representations to become
increasingly similar, reducing the ability of the network to preserve
node-specific information. Second, because Graph4BiLO is embedded directly
inside a mixed-integer optimization model, every additional message-passing
layer introduces another collection of node-wise ReLU activations and their
associated mixed-integer constraints. Thus, increasing depth simultaneously
increases representational reach and the computational burden of the embedded
model.

To examine the first effect at the depth required for complete local graph
coverage, we train a four-layer Graph4BiLO model with hidden dimension
\(d=4\) on \(n=20\) knapsack-interdiction instances. For each held-out test
graph and each message-passing layer, we compute the cosine similarity between
all distinct pairs of nodes of the same type and average the off-diagonal
values. Similarity approaching one indicates that different nodes have
developed nearly parallel representations and therefore retain less
node-specific distinction.

\begin{table*}[t]
\centering
\caption{Mean within-type pairwise cosine similarity across message-passing
depth for a four-layer Graph4BiLO model with hidden dimension \(d=4\),
trained and evaluated on \(n=20\) knapsack-interdiction instances. Layer 4
corresponds to the depth required for complete local receptive coverage of
the graph.}
\label{tab:cosine_similarity}
\begin{tabular}{c c c c c c c}
\toprule
Node Type &
Layer 0 &
Layer 1 &
Layer 2 &
Layer 3 &
Layer 4 &
Post-Broadcast \\
\midrule
Follower-variable ($Y$)
& 0.931 & 0.955 & 0.892 & 0.689 & 0.947 & 0.985 \\

Constraint ($c$)
& 0.935 & 0.889 & 0.701 & 0.890 & 0.970 & 1.000 \\

Leader-variable ($x$)
& 0.736 & 0.593 & 0.991 & 0.650 & 0.986 & 0.997 \\
\bottomrule
\end{tabular}
\end{table*}

Table~\ref{tab:cosine_similarity} shows that by the fourth message-passing
layer, when information can propagate across the full graph, representations
within each node type have become highly similar. Mean pairwise cosine
similarity reaches \(0.947\) for follower-variable nodes, \(0.970\) for
constraint nodes, and \(0.986\) for leader-variable nodes. The progression is
not monotonic across intermediate layers, so the results do not imply that
every additional layer necessarily increases similarity. Rather, they show the depth required for complete local graph coverage
coincides with substantial representation homogenization.

The subsequent global broadcast increases the similarities further, to
\(0.985\), \(1.000\), and \(0.997\), respectively. This is expected because
the same graph-level summary is supplied to every node. The broadcast is
useful precisely because it provides global context without requiring
additional local message-passing layers, but it also illustrates the tension
between sharing global information and preserving distinct local
representations.

The second cost of increasing depth is computational. In the embedded
formulation, every additional message-passing layer creates another
\(d\)-dimensional ReLU representation for every graph node. The architecture
ablation in Section~\ref{sec:architecture_ablation} confirms that deeper
models are substantially more expensive to optimize: on \(n=20\) instances
with \(d=4\), the one-layer model solves in \(0.28\) seconds on average,
whereas the two-, three-, and four-layer models require \(3.20\), \(4.30\),
and \(3.49\) seconds, respectively.

These results expose a central limitation of using local message-passing GNNs
inside optimization models. Capturing the full structure of an optimization
problem may require a receptive field spanning the complete graph, which in
this case requires four message-passing layers. Yet increasing depth to obtain
that coverage can both homogenize node representations and enlarge the
embedded mixed-integer formulation. Graph4BiLO must therefore balance the
benefit of broader information propagation against two costs: loss of
node-level distinction and increased optimization complexity.

For this reason, the main Graph4BiLO configuration (Table~\ref{tab:performance}) uses two local
message-passing layers followed by a global pooling-and-broadcast operation.
The two message-passing layers provide local multi-hop interaction without
incurring the full depth required for complete graph traversal, while the
broadcast step supplies every node with a summary of the entire optimization
instance. This design is intended to balance local structural reasoning,
global information access, and the computational cost of the embedded GNN.

\section{Discussion}

\subsection{Practical Guidance}

The results suggest that the choice between Graph4BiLO and a more compact
learning-based surrogate should depend primarily on how the model will be
used. When optimization repeatedly occurs at a single fixed problem size,
a compact surrogate such as Neur2BiLO is attractive: on the present
knapsack-interdiction benchmark it achieves essentially the same objective
quality as Graph4BiLO while producing a much smaller and faster embedded
optimization model. Similarly, when instances are sufficiently small and
certified optimality is important, an exact bilevel solver remains preferable
when its computational cost is acceptable.

Graph4BiLO becomes more attractive when a common model must be reused across
a family of problem sizes. Its shared message-passing parameters are independent
of the number of graph nodes, and the OOD experiment demonstrates that a model
trained exclusively at \(n=20\) can be applied directly to previously unseen
\(n=40\) and \(n=60\) instances without retraining or fine-tuning. This
property is especially relevant when generating supervised targets is costly,
since every label requires an exact lower-level solve. Under a size-specific
training protocol, supporting \(n=20\), \(40\), and \(60\) with 2,000
training examples per size would require 6,000 lower-level solves and three
training procedures; the OOD Graph4BiLO experiment uses 2,000 labeled
\(n=20\) instances and a single trained model. The potential savings become
more important when the follower itself is a difficult combinatorial
optimization problem, as in traveling-salesman and vehicle-routing variants
\cite{papadimitriou1977euclidean,lenstra1981complexity}.

The practical tradeoff is therefore not simply predictive accuracy versus
computational efficiency. Graph4BiLO exchanges a substantially larger embedded
MILP for structural parameter sharing and demonstrated cross-size transfer.
A practitioner solving one fixed problem dimension may reasonably prefer the
more compact representation, whereas applications involving changing problem
dimensions, expensive label generation, or optimization structures whose
connectivity carries important information provide a stronger motivation for
the graph-based formulation. This tradeoff can also be moderated by selecting
a smaller hidden dimension \(d\), which reduces the number of embedded ReLU
activations and therefore the size of the resulting mixed-integer formulation.

\subsection{Limitations and Future Directions}

Several limitations define the scope of the present results. First, the
cross-size experiment establishes zero-shot transfer for Graph4BiLO but does
not establish that Graph4BiLO has superior cross-size generalization to every
other variable-size neural architecture. In particular, although the reported
Neur2BiLO knapsack experiments use size-specific training and checkpoints
\cite{dumouchelle2024neur2bilo,neur2bilo_github}, its set-based architecture
is in principle compatible with variable-sized inputs. A direct comparison
between alternative architectures trained under the same cross-size protocol
would therefore be valuable future work.

Second, the present evaluation is limited to knapsack interdiction, and generalizing the results to other bilevel problem classes is an important direction for future work.

Third, embedding the GNN directly within the optimization model remains the
principal scalability limitation. Every node-wise ReLU activation can
introduce additional continuous variables, binary activation indicators, and
linear constraints, so the size of the embedded formulation grows with the
number of graph nodes, hidden dimension, and message-passing depth. The
architecture ablation further shows that increasing depth does not
monotonically improve prediction quality, while the representation-similarity
analysis indicates substantial homogenization at the full four-hop depth. Future work should therefore investigate
whether the mixed-integer representation of the trained GNN can be reduced
without altering the underlying network or sacrificing its predictive and
cross-size generalization benefits.

Finally, approximation error interacts directly with feasibility of the
learned value-function constraint. The current slack variable prevents
overprediction from rendering the surrogate model infeasible, and the final
exact follower solve restores lower-level optimality for the selected leader
decision. Future work should investigate uncertainty-aware, conservative, or
one-sided value-function learning objectives that explicitly account for the
different optimization consequences of underprediction and overprediction.

\section{Conclusion}
This paper introduced Graph4BiLO, a graph neural network framework for approximating value functions in bilevel mixed-integer linear optimization. Graph4BiLO represents optimization instances as variable--constraint graphs, learns the follower value function using shared message-passing parameters, and embeds the trained ReLU network directly in a single-level mixed-integer formulation. A final exact follower solve repairs the learned solution and restores lower-level optimality.

On knapsack interdiction instances with 20--100 items, Graph4BiLO produced objective values comparable to Neur2BiLO while using one trained model across multiple problem sizes. The primary limitation was optimization time: the node-wise ReLU representation creates a large number of additional binary and continuous variables, causing the embedded MILP to become difficult to solve for larger instances. The results therefore suggest that GNNs are a promising structural
representation for learned bilevel value functions, while highlighting the
computational cost of their mixed-integer embeddings as the primary remaining
scalability challenge.

\bibliographystyle{IEEEtran}
\bibliography{references}

@inproceedings{dumouchelle2024neur2bilo,
  author    = {Justin Dumouchelle and Esther Julien and Jannis Kurtz and Elias B. Khalil},
  title     = {{Neur2BiLO}: Neural Bilevel Optimization},
  booktitle = {Advances in Neural Information Processing Systems},
  year      = {2024},
  doi       = {10.52202/079017-2752}
}

@article{tahernejad2020mibs,
  author  = {Sahar Tahernejad and Ted K. Ralphs and Scott T. DeNegre},
  title   = {A Branch-and-Cut Algorithm for Mixed Integer Bilevel Linear Optimization Problems and Its Implementation},
  journal = {Mathematical Programming Computation},
  volume  = {12},
  number  = {4},
  pages   = {529--568},
  year    = {2020},
  doi     = {10.1007/s12532-020-00183-6}
}

@inproceedings{schlichtkrull2018modeling,
  title={Modeling relational data with graph convolutional networks},
  author={Schlichtkrull, Michael and Kipf, Thomas N and Bloem, Peter and Van Den Berg, Rianne and Titov, Ivan and Welling, Max},
  booktitle={European semantic web conference},
  pages={593--607},
  year={2018},
  organization={Springer}
}

@article{vasquez2025single,
  title={A single-level reformulation of binary bilevel programs using decision diagrams: S. V{\'a}squez et al.},
  author={V{\'a}squez, Sebasti{\'a}n and Lozano, Leonardo and van Hoeve, Willem-Jan},
  journal={Mathematical Programming},
  pages={1--54},
  year={2025},
  publisher={Springer}
}

@book{dempe2002foundations,
  title={Foundations of bilevel programming},
  author={Dempe, Stephan},
  year={2002},
  publisher={Springer}
}

@article{bard1998practical,
  title={Practical bilevel optimization},
  author={Bard, Jonathan F},
  journal={The Netherlands: Kluwer Academic Publishers},
  year={1998}
}

@article{dempe2007new,
  title={New necessary optimality conditions in optimistic bilevel programming},
  author={Dempe, Stephan and Dutta, Joydeep and Mordukhovich, BS},
  journal={Optimization},
  volume={56},
  number={5-6},
  pages={577--604},
  year={2007},
  publisher={Taylor \& Francis}
}

@inproceedings{zhou2024learning,
  title={Learning to solve bilevel programs with binary tender},
  author={Zhou, Bo and Jiang, Ruiwei and Shen, Siqian},
  booktitle={International Conference on Learning Representations},
  volume={2024},
  pages={31886--31908},
  year={2024}
}

@article{kwon2025deep,
  title={Deep learning based high accuracy heuristic approach for knapsack interdiction problem},
  author={Kwon, Sunhyeon and Choi, Hwayong and Park, Sungsoo},
  journal={Computers \& Operations Research},
  volume={176},
  pages={106965},
  year={2025},
  publisher={Elsevier}
}

@inproceedings{networkinterdictiongoesneural,
author = {Zhang, Lei and Chen, Zhiqian and Lu, Chang-Tien and Zhao, Liang},
title = {Network Interdiction Goes Neural},
year = {2025},
isbn = {9798400714542},
publisher = {Association for Computing Machinery},
address = {New York, NY, USA},
url = {https://doi.org/10.1145/3711896.3737063},
doi = {10.1145/3711896.3737063},
booktitle = {Proceedings of the 31st ACM SIGKDD Conference on Knowledge Discovery and Data Mining V.2},
pages = {3774–3785},
numpages = {12},
location = {Toronto ON, Canada},
series = {KDD '25}
}

@article{gasse2019exact,
  title={Exact combinatorial optimization with graph convolutional neural networks},
  author={Gasse, Maxime and Ch{\'e}telat, Didier and Ferroni, Nicola and Charlin, Laurent and Lodi, Andrea},
  journal={Advances in neural information processing systems},
  volume={32},
  year={2019}
}

@article{han2023gnn,
  title={A gnn-guided predict-and-search framework for mixed-integer linear programming},
  author={Han, Qingyu and Yang, Linxin and Chen, Qian and Zhou, Xiang and Zhang, Dong and Wang, Akang and Sun, Ruoyu and Luo, Xiaodong},
  journal={arXiv preprint arXiv:2302.05636},
  year={2023}
}

@article{canturk2024scalable,
  title={Scalable primal heuristics using graph neural networks for combinatorial optimization},
  author={Cant{\"u}rk, Furkan and Varol, Taha and Aydo{\u{g}}an, Reyhan and {\"O}zener, Okan {\"O}rsan},
  journal={Journal of Artificial Intelligence Research},
  volume={80},
  pages={327--376},
  year={2024}
}

@article{fischetti2018deep,
  title={Deep neural networks and mixed integer linear optimization},
  author={Fischetti, Matteo and Jo, Jason},
  journal={Constraints},
  volume={23},
  number={3},
  pages={296--309},
  year={2018},
  publisher={Springer}
}

@article{anderson2020strong,
  title={Strong mixed-integer programming formulations for trained neural networks},
  author={Anderson, Ross and Huchette, Joey and Ma, Will and Tjandraatmadja, Christian and Vielma, Juan Pablo},
  journal={Mathematical Programming},
  volume={183},
  number={1},
  pages={3--39},
  year={2020},
  publisher={Springer}
}

@article{zhang2024augmenting,
  title={Augmenting optimization-based molecular design with graph neural networks},
  author={Zhang, Shiqiang and Campos, Juan S and Feldmann, Christian and Sandfort, Frederik and Mathea, Miriam and Misener, Ruth},
  journal={Computers \& Chemical Engineering},
  volume={186},
  pages={108684},
  year={2024},
  publisher={Elsevier}
}

@article{jeroslow1985polynomial,
  title={The polynomial hierarchy and a simple model for competitive analysis},
  author={Jeroslow, Robert G},
  journal={Mathematical programming},
  volume={32},
  number={2},
  pages={146--164},
  year={1985},
  publisher={Springer}
}

@inproceedings{karp1972reducibility,
  title={Reducibility among combinatorial problems},
  author={Karp, Richard M},
  booktitle={Complexity of Computer Computations: Proceedings of a symposium on the Complexity of Computer Computations, held March 20--22, 1972, at the IBM Thomas J. Watson Research Center, Yorktown Heights, New York, and sponsored by the Office of Naval Research, Mathematics Program, IBM World Trade Corporation, and the IBM Research Mathematical Sciences Department},
  pages={85--103},
  year={1972},
  organization={Springer}
}

@article{fischetti2019interdiction,
  title={Interdiction games and monotonicity, with application to knapsack problems},
  author={Fischetti, Matteo and Ljubi{\'c}, Ivana and Monaci, Michele and Sinnl, Markus},
  journal={INFORMS Journal on Computing},
  volume={31},
  number={2},
  pages={390--410},
  year={2019},
  publisher={INFORMS}
}

@article{caprara2016bilevel,
  title={Bilevel knapsack with interdiction constraints},
  author={Caprara, Alberto and Carvalho, Margarida and Lodi, Andrea and Woeginger, Gerhard J},
  journal={INFORMS Journal on Computing},
  volume={28},
  number={2},
  pages={319--333},
  year={2016},
  publisher={INFORMS}
}

@article{caprara2014study,
  title={A study on the computational complexity of the bilevel knapsack problem},
  author={Caprara, Alberto and Carvalho, Margarida and Lodi, Andrea and Woeginger, Gerhard J},
  journal={SIAM Journal on Optimization},
  volume={24},
  number={2},
  pages={823--838},
  year={2014},
  publisher={SIAM}
}

@article{papadimitriou1977euclidean,
  author  = {Christos H. Papadimitriou},
  title   = {The Euclidean Travelling Salesman Problem is NP-Complete},
  journal = {Theoretical Computer Science},
  volume  = {4},
  number  = {3},
  pages   = {237--244},
  year    = {1977},
  doi     = {10.1016/0304-3975(77)90012-3}
}

@article{lenstra1981complexity,
  author  = {Jan Karel Lenstra and Alexander H. G. Rinnooy Kan},
  title   = {Complexity of Vehicle Routing and Scheduling Problems},
  journal = {Networks},
  volume  = {11},
  number  = {2},
  pages   = {221--227},
  year    = {1981},
  doi     = {10.1002/net.3230110211}
}

@misc{neur2bilo_github,
  author       = {{Khalil Research}},
  title        = {{Neur2BiLO}: Neural Bilevel Optimization},
  howpublished = {\url{https://github.com/khalil-research/Neur2BiLO}},
  note         = {GitHub repository, accessed August 21, 2026},
  year         = {2024}
}

@inproceedings{fey2019pytorch,
  title     = {Fast Graph Representation Learning with {PyTorch Geometric}},
  author    = {Fey, Matthias and Lenssen, Jan Eric},
  booktitle = {ICLR Workshop on Representation Learning on Graphs and Manifolds},
  year      = {2019}
}

\end{document}